\documentclass{article}

\PassOptionsToPackage{numbers, compress}{natbib}

\usepackage[preprint]{neurips_2022}

\usepackage[utf8]{inputenc} 
\usepackage[T1]{fontenc}    
\usepackage{hyperref}       
\usepackage{url}            
\usepackage{booktabs}       
\usepackage{amsfonts}   
\usepackage{nicefrac}       
\usepackage{microtype}      
\usepackage{xcolor}         
\usepackage{natbib}
\usepackage{adjustbox}
\usepackage{multirow}
\usepackage{pgfplots}
\usepackage{subcaption}
\usepackage{amsmath}
\pgfplotsset{compat=1.18}

\usepackage{graphics} 
\title{Anomaly detection using Diffusion-based methods}

\author{%
  Aryan Bhosale \\
  Indian Institute of Technology Bombay \\
  Mumbai, India \\
  \And
  Samrat Mukherjee \\
  Indian Institute of Technology Bombay \\
  Mumbai, India \\
  \And
  Biplab Banerjee \\
  Indian Institute of Technology Bombay \\
  Mumbai, India \\
  \And
  Fabio Cuzzolin \\
  Oxford-Brookes University \\
  Oxford, UK \\
}

\begin{document}

\maketitle

\begin{abstract}
This paper explores the utility of diffusion-based models for anomaly detection, focusing on their efficacy in identifying deviations in both compact and high-resolution datasets. Diffusion-based architectures, including Denoising Diffusion Probabilistic Models (DDPMs) and Diffusion Transformers (DiTs), are evaluated for their performance using reconstruction objectives. By leveraging the strengths of these models, this study benchmarks their performance against traditional anomaly detection methods such as Isolation Forests, One-Class SVMs, and COPOD. The results demonstrate the superior adaptability, scalability, and robustness of diffusion-based methods in handling complex real-world anomaly detection tasks. Key findings highlight the role of reconstruction error in enhancing detection accuracy and underscore the scalability of these models to high-dimensional datasets. Future directions include optimizing encoder-decoder architectures and exploring multi-modal datasets to further advance diffusion-based anomaly detection.
\end{abstract}

\section{Introduction}

Anomaly detection (AD) is a fundamental problem in machine learning with applications in domains such as industrial fault detection, healthcare diagnostics, and cybersecurity. It focuses on identifying rare and often subtle deviations from expected patterns within data. Traditional methods, including Isolation Forest (IForest) \cite{liu2008isolating}, One-Class Support Vector Machine (OCSVM) \cite{scholkopf2001estimating}, and COPOD \cite{li2020copod}, are widely employed for this task. However, these methods face significant challenges in handling complex, high-dimensional, and noisy datasets, particularly in real-world scenarios where data distributions can evolve over time or include nuanced anomalies.

Recent advancements in generative models, particularly diffusion models, provide a promising new avenue for addressing these challenges. Diffusion models such as Denoising Diffusion Probabilistic Models (DDPMs) \cite{ho2020denoising}, Diffusion Transformers (DiTs) \cite{song2021denoising}, and Denoising Transformer Embeddings (DTE) \cite{song2021denoising} are designed to iteratively reconstruct data from noise, capturing intricate patterns in high-dimensional spaces. These models excel at modeling complex data distributions, making them particularly suitable for anomaly detection in challenging settings, including noisy and adversarial environments.

In this work, we systematically evaluate the efficacy of diffusion-based methods for anomaly detection across a diverse range of datasets, comparing their performance to traditional techniques. Using the Area Under the Receiver Operating Characteristic (AUC-ROC) as the evaluation metric, we demonstrate that diffusion models offer significant advantages in terms of adaptability, scalability, and robustness.

\section{Related Works}

\subsection{Anomaly Detection}

Villa-Perez \textit{et al}. \cite{villa2021semi}, analyze 29 state-of-the-art SSAD algorithms for anomaly detection. Amongst these are methods based on K Nearest Neighbours (KNN), GANs, VAEs, isolation forests and ensemble based methods. Bagging-Random Miner acheived state-of-the-art results when tested across 95 different datasets and the average AUC  was taken. This method is an ensemble based method which is domain-specific in masquerade detection. \cite{camina2019bagging}. These algorithms include one class SVMS (ocSVM), isolation forests, and KNNs. Villa et al. \cite{villa2021semi} proposed a semi-supervised anomaly detection framework that combines self-supervised pretraining with feature refinement, enhancing detection accuracy on diverse datasets.

The field of deep anomaly detection has grown significantly in recent years. \cite{pang2021deep} provided a comprehensive survey of methods, emphasizing the benefits of deep learning over traditional approaches. \cite{ruff2021unifying} extended this by reviewing both shallow and deep methods, highlighting advancements in learning tailored representations for anomaly detection. Key deep learning models include DeepSVDD, which utilizes a one-class objective for representation learning \cite{ruff2018deep}, and DAGMM, which integrates deep learning with Gaussian Mixture Models for anomaly detection \cite{zong2018deep}. Contrastive learning methods like ICL \cite{shenkar2022icl} have also emerged as powerful tools for semi-supervised anomaly detection, achieving competitive results on ODDS datasets.

Several notable works have extended anomaly detection methodologies beyond autoencoders, VAEs, and GANs. Robust Deep Autoencoders \cite{zhou2017anomaly} enhance traditional autoencoder structures with sparsity constraints for anomaly detection. DROCC \cite{goyal2020drocc} introduces a contrastive optimization framework, while GOAD \cite{bergman2020classification} employs geometric transformations for one-class classification. Lunar \cite{goodge2022lunar} is another promising approach focusing on high-dimensional anomaly detection.

As a baseline method put forward by  \cite{fabiocssl}, self-training uses the model to create pseudo-labels which, if confident enough, are accepted as ground truth labels to enable the model to train in a supervised way - thereby reducing the problem to a continual supervised learning problem. Alternatively, conditional Triple-GANs were used by \cite{wang2021ordisco} in order to accept optionally labelled data using a classifier, generator and discriminator to model the joint distribution of the data for use in generative replay. This proved incredibly effective, albeit with few comparisons beside memory buffers which are often seen as relaxing the constraints of continual learning, rather than directly solving the problem. 

\subsection{VAE-Based Approaches}
Variational Autoencoders (VAEs) have been widely adopted for anomaly detection due to their ability to model data distributions and detect deviations. \cite{an2015variational} introduced one of the first VAE-based approaches, leveraging reconstruction probabilities to identify anomalies. \cite{sakurada2014anomaly} explored the potential of autoencoders for time-series anomaly detection, while \cite{xia2015learning} combined deep learning with domain knowledge to enhance detection performance. These methods underscore the flexibility of VAEs in various anomaly detection settings.

\subsection{GAN-Based Approaches}
Generative Adversarial Networks (GANs) have gained traction for anomaly detection by generating synthetic data to augment training. SO-GAAL and MO-GAAL \cite{liu2019generative} apply GAN frameworks to generate anomalies, enabling more robust learning for rare-event detection. These methods focus on balancing the adversarial objectives of the generator and discriminator to improve anomaly detection accuracy. GANs have proven particularly effective in scenarios with limited labeled data.

\subsection{Diffusion-Based Techniques}
Diffusion models have recently emerged as an innovative approach for anomaly detection. While initially applied to image-based one-class settings, recent efforts have expanded their use to tabular data and unsupervised scenarios. \cite{wolleb2022diffusion} proposed a method using a diffusion process followed by classifier-guided denoising. \cite{zhang2023diffusion} synthesized anomalous samples to train denoising networks for anomaly repair. AnoDDPM \cite{wyatt2022anoddpm} trains a denoising network for normal image reconstruction using diffusion noise, while \cite{graham2023ddpm} combined DDPM-based reconstructions across timesteps to compute anomaly scores. \cite{liu2023checkerboard} introduced a diffusion approach that reconstructs images via in-painting based on checkerboard-masked inputs, showcasing the potential of diffusion models in this domain.

\section{Our Approach}
\subsection{\textbf{Preliminaries}}

This section reviews the concepts behind Denoising Diffusion Probabilistic Models (DDPMs) as described by Dickstein et. al \cite{sohl2015thermodynamics, ho2020ddpm}. DDPMs are generative models that approximate complex data distributions by simulating a diffusion process. This process can be viewed as a sequence of steps where data is gradually corrupted by noise and then reconstructed by learning the reverse of this noising process. This framework is formalized using Markov chains, which can accurately capture the diffusion and reverse diffusion processes.

\subsection{The Forward Diffusion Process}
Consider a dataset with real data instances \( x_0 \in \mathbb{R}^d \), where the data is sampled from an unknown distribution \( q( x_0) \). The forward diffusion process is defined as a discrete-time Markov chain that progressively adds noise to the data through a series of \( T \) timesteps. 

At each timestep \( t \), a noisy version of the data \( x_t \) is generated conditioned on the previous state \( x_{t-1} \), according to the conditional distribution:

\[
q(x_t | x_{t-1}) := \mathcal{N}(x_t; \sqrt{1 - \beta_t} x_{t-1}, \beta_t \mathbf{I}),
\]
where \( \beta_t \in [0, 1] \) controls the variance of the noise added at each timestep. The transition from \( x_0 \) to \( x_T \), where \( x_T \) represents a noisy sample at the final timestep, is governed by the marginal distribution:

\[
q(x_t | x_0) := \mathcal{N}(x_t; \sqrt{\bar{\alpha}_t} x_0, (1 - \bar{\alpha}_t) \mathbf{I}),
\]
where \( \alpha_t := 1 - \beta_t \) and \( \bar{\alpha}_t := \prod_{s=1}^t \alpha_s \) is the cumulative product over the timesteps. 
In practice, the forward process is implemented by sampling from a Gaussian distribution at each timestep using the reparameterization trick. Specifically, the noisy state \( x_t \) at timestep \( t \) is given by:

\[
x_t = \sqrt{\bar{\alpha}_t} x_0 + \sqrt{1 - \bar{\alpha}_t} \epsilon_t,
\]
where \( \epsilon_t \sim \mathcal{N}(0, \mathbf{I}) \) is standard Gaussian noise. The forward diffusion process can be viewed as a Markov chain \( \{x_t\}_{t=0}^{T} \) where the transition at each timestep depends solely on the previous state.

\subsection{Denoising Process}
The reverse diffusion process models the denoising operation by progressively removing noise from \( x_T \) (which is initially pure noise) to recover the data \(x_0 \). The reverse process is also a Markov chain, and it is parameterized by a neural network \( p_\theta(x_{t-1} | x_t) \), which learns the transition dynamics of the reverse process. The reverse conditional distribution \( p_\theta(x_{t-1} | x_t) \) is parameterized by a neural network and is typically modeled as a Gaussian:

\[
p_\theta(x_{t-1} | x_t) := \mathcal{N}(x_{t-1}; \boldsymbol{\mu}_\theta(x_t, t), \boldsymbol{\Sigma}_\theta(x_t, t)),
\]
where \( \boldsymbol{\mu}_\theta(x_t, t) \) and \( \boldsymbol{\Sigma}_\theta(x_t, t) \) are the predicted mean and covariance of the reverse transition, both of which are learned during training. In the reverse diffusion process, neural networks such as U-Nets \cite{ronneberger2015u} and Vision Transformers (ViTs) \cite{dosovitskiy2020discriminative} are commonly used. U-Nets, with their encoder-decoder structure and skip connections, excel at preserving local spatial details, making them well-suited for denoising tasks. ViTs, leveraging self-attention, capture global dependencies and are more effective in handling complex, high-resolution image generation tasks. In this work, we focus on Diffusion Transformers, which utilize ViTs for the reverse diffusion process, leveraging their ability to model long-range dependencies and generate high-quality images in diffusion models \cite{dhariwal2021diffusiontransformer}.

\subsection{Training Diffusion Models}

The model is trained by optimizing the variational lower bound of the log-likelihood of \( x_0 \) \cite{kingma2013auto}, which simplifies to:

\[
\mathcal{L}(\theta) = -p(x_0 | x_1) + \sum_t \mathcal{D}_{KL}\left( q(x_{t-1} | x_t, x_0) \ \Big\| \ p_\theta(x_{t-1} | x_t) \right),
\]
where the Kullback-Leibler divergence \( \mathcal{D}_{KL} \) is computed using the mean and covariance of both distributions. To enable training, the mean \( \mu_\theta \) is reparameterized as a noise prediction network \( \epsilon_\theta \), and the model is trained with the mean squared error loss between the predicted noise and the noise added in the forward diffusion process:

\[
\mathcal{L}_{simple}(\theta) = \left\| \epsilon_\theta(x_t) - \epsilon_t \right\|_2^2,
\]
where \( \epsilon_t \) is the ground truth noise and \(\epsilon_\theta(x_t)\) is the predicted noise. We follow the approach of Nichol and Dhariwal \cite{nichol2021improved}, where \( \epsilon_\theta \) is trained using \( \mathcal{L}_{simple} \). We optimize the reverse process covariance \( \Sigma_\theta(x_t) \) using the full loss function, which incorporates the Kullback-Leibler divergence between the approximate posterior and the reverse process:

\[
\mathcal{L}(\theta) = \mathbb{E}_{q(x_t, x_0)} \left[ \mathcal{D}_{KL} \left( q^*(x_{t-1} | x_t, x_0) \ \Big\| \ p_\theta(x_{t-1} | x_t) \right) \right].
\]

Once trained, new samples are generated by initializing \( x_{t_\text{max}} \sim \mathcal{N}(0, \mathbf{I}) \) and sampling iteratively from:

\[
x_{t-1} \sim \mathcal{N}\left( x_{t-1}; \boldsymbol{\mu}_\theta(x_t, t), \boldsymbol{\Sigma}_\theta(x_t, t) \right),
\]
where \( \boldsymbol{\mu}_\theta(x_t, t) \) and \( \boldsymbol{\Sigma}_\theta(x_t, t) \) are the mean and covariance predicted by the model. This reverse process is performed via the reparameterization trick, which allows sampling from the learned reverse diffusion process \cite{dhariwal2021diffusiontransformer}.

\subsubsection*{Reconstruction Error}
During training, Diffusion models learn to model the normal distribution by learning a forward (adding noise) and reverse process (denoising). Reconstruction error, measured as the difference between the input and the denoised reconstruction of the same via the Diffusion model is a measure to distinguish between normal and anomalous examples \cite{he2023diad}. Since anomalies are absent or underrepresented, the diffusion model fails to learn a trajectory to faithfully learn the reconstruction process and hence results in higher error for such samples as seen in \cite{wolleb2022diffusion, he2023diad, livernoche2023diffusion}.

\subsection{Methodology followed}

In this paper, we explore the capabilities of diffusion-based architectures to tackle anomaly detection tasks formulated using the reconstruction objective. Diffusion models, originally designed for generative tasks, learn to iteratively refine samples by denoising through a sequence of time steps. We specifically evaluate two prominent diffusion-based architectures: Denoising Diffusion Probabilistic Models (DDPMs) and Diffusion Transformers (DiTs). These architectures offer complementary advantages.

Denoising Diffusion Probabilistic Models (DDPMs) utilize a U-Net architecture to effectively model the denoising process, making them ideal for anomaly detection. The hierarchical structure of U-Net captures intricate spatial details through skip connections while maintaining global context, allowing for precise reconstruction of images. This capability enables the identification of anomalies by highlighting discrepancies between the original and reconstructed outputs, thus enhancing detection accuracy.

Diffusion-based Transformers (DiTs) leverage the Vision Transformer (ViT) architecture to enhance the diffusion process for high-resolution image generation. By processing image patches as input tokens and utilizing self-attention mechanisms, DiTs effectively capture global dependencies and multi-scale features, making them adept at handling complex visual datasets such as ImageNet. This innovative approach not only improves scalability but also significantly enhances performance in detecting anomalies across diverse visual domains.

High-resolution datasets like ImageNet pose significant challenges for anomaly detection due to their complex spatial details and extensive feature spaces. Transformer-based architectures, such as DiTs, effectively address these challenges by enabling efficient parallel processing and adaptive focus on relevant image regions, enhancing their capability to detect subtle anomalies in high-resolution data.

Our experiments reveal that diffusion-based architectures, especially DiTs, excel in scalable high-resolution anomaly detection.

\section{Results}

\subsection{Datasets}

We have performed experiments on a diverse collection of datasets to benchmark anomaly detection methods. We conducted experiments on two categories of datasets:

\begin{itemize}
    \item \textbf{Compact Datasets:} This category includes datasets from ADBench, such as CIFAR-10, MNIST-C, and MVTec-AD, which consist of small-scale images (e.g., 28×28 or 32×32 pixels). These datasets collectively provide a comprehensive evaluation of anomaly detection methods, covering noise-based anomalies (e.g., MNIST-C), semantic out-of-distribution challenges (e.g., CIFAR-10 subsets), and real-world industrial defect detection (e.g., MVTec-AD).
    \item \textbf{High-Resolution Datasets:} This category is represented by subsets of the ImageNet (Mini-ImageNet) dataset, consisting of high-resolution images (224×224 pixels). 
\end{itemize}

\subsection{Compact Datasets (ADBench)}
Compact datasets primarily feature small-scale images, making them ideal for benchmarking anomaly detection methods in low-resolution scenarios. These datasets include:
\begin{itemize}
    \item \textbf{CIFAR-10:} CIFAR-10 is a benchmark dataset comprising 60,000 32×32 color images across 10 classes. We utilize the following subsets:
        \begin{itemize}
            \item \textbf{CIFAR10\_x}: This notation represents various experimental splits of the CIFAR-10 dataset, where \( x \) denotes the specific subset number. In these subsets, class \( x \) samples are treated as anomalies (out-of-distribution), while the remaining classes are utilized for training. This setup effectively simulates out-of-distribution detection tasks, allowing researchers to evaluate model performance in identifying and handling anomalies within a controlled environment.
        \end{itemize}
    \item \textbf{MNIST-Corrupted (MNIST-C):} MNIST-C contains 28×28 grayscale images and extends the original MNIST dataset with systematic corruptions. The corruptions used include:
        \begin{itemize}
            \item \textbf{MNIST-C\_spatter}: Adds splatter-like noise to images.
            \item \textbf{MNIST-C\_dotted\_line}: Introduces dotted lines, mimicking degraded handwriting.
            \item \textbf{MNIST-C\_shot\_noise}: Applies shot noise, simulating errors during image acquisition.
            \item \textbf{MNIST-C\_shear}: Applies geometric shearing transformations.
            \item \textbf{MNIST-C\_fog}: Adds fog-like artifacts, reducing visibility.
        \end{itemize}
    \item \textbf{Street View House Numbers (SVHN)}: SVHN contains 32×32 digit images from real-world house numbers. We use the following subsets:
        \begin{itemize}
            \item  \textbf{SVHN\_x}: In these subsets, class \( x \) samples are designated as anomalous samples while others are classified as normal.
        \end{itemize}
    \item \textbf{MVTec-AD:} MVTec-AD is a high-resolution industrial dataset with normal and defective samples. It includes images of industrial objects and textures, focusing on detecting structural defects. Categories used include:
        \begin{itemize}
            \item \textbf{MVTec-AD\_category}: Each category includes normal samples and anomalies such as scratches, dents, missing components, or irregular textures. Category can be tile, grid, bottle, capsule, cable, carpet, leather, metal\_nut, pill, zipper. 
        \end{itemize}
\end{itemize}

\subsection{High-Resolution Datasets (ImageNet Subsets)}
High-resolution datasets provide a more challenging benchmark for anomaly detection. ImageNet, with its diverse categories and high-resolution images, introduces additional complexity in terms of spatial detail and feature richness. They challenge the model to scale effectively while retaining performance. This is particularly important for diffusion-based transformer (DiT) models, which are designed to capture long-range dependencies and hierarchical structures. This helped us to validate scalability by transitioning from low-resolution to high-resolution datasets and to motivate real-world applications of DiTs in practical scenarios.

\subsection{Experimental Results}

\begin{table}[ht]
    \centering
    
    \begin{adjustbox}{width=\textwidth}
    \begin{tabular}{lcccccc}
        \toprule
        \textbf{Dataset} & \textbf{IForest} & \textbf{OCSVM} & \textbf{COPOD} & \textbf{DDPM} & \textbf{DiT} & \textbf{DTE} \\
        \midrule
        CIFAR10\_5 & 53.66 & \underline{58.62} & 46.32 & 58.19 & 58.12 & \textbf{59.41} \\
        MNIST-C\_spatter & 85.13 & 86.20 & 50.00 & \underline{86.65} & 84.94 & \textbf{89.94} \\
        MNIST-C\_dotted\_line & 78.70 & 81.18 & 50.00 & \textbf{82.01} & \underline{81.62} & 52.92 \\
        SVHN\_5 & 60.17 & 63.09 & 50.00 & \underline{63.29} & \textbf{63.75} & 61.26 \\
        CIFAR10\_6 & \underline{72.14} & 71.37 & 67.41 & \textbf{72.26} & 71.45 & 53.39 \\
        MVTec-AD\_tile & 83.82 & 84.75 & 50.00 & \underline{85.59} & 85.27 & \textbf{92.74} \\
        FashionMNIST\_6 & 67.13 & 74.34 & 50.00 & \underline{75.16} & 73.18 & \textbf{77.90} \\
        MVTec-AD\_zipper & \underline{82.12} & 80.62 & 50.00 & \textbf{83.21} & 79.05 & 49.08 \\
        MVTec-AD\_grid & 63.67 & \underline{65.35} & 50.00 & \textbf{66.94} & 64.80 & 51.22 \\
        SVHN\_3 & 56.31 & 59.56 & 50.00 & \underline{59.86} & \textbf{60.20} & 53.11 \\
        CIFAR10\_9 & 69.04 & \underline{72.54} & 50.00 & \textbf{72.60} & 71.97 & 56.01 \\
        MVTec-AD\_bottle & \textbf{96.88} & 96.53 & 50.00 & \underline{96.83} & 96.68 & 50.23 \\
        SVHN\_7 & \underline{67.47} & 65.79 & 50.00 & 67.16 & \textbf{67.21} & 63.50 \\
        MVTec-AD\_carpet & \underline{70.35} & 68.42 & 50.00 & 69.89 & 69.07 & \textbf{82.36} \\
        MVTec-AD\_metal\_nut & 71.97 & \underline{73.51} & 50.00 & \textbf{76.67} & 71.57 & 54.28 \\
        CIFAR10\_4 & 76.31 & \underline{76.93} & 50.00 & \textbf{77.42} & 77.07 & 55.40 \\
        MNIST-C\_shot\_noise & 80.55 & \underline{81.28} & 50.00 & \textbf{82.15} & 79.75 & 54.10 \\
        FashionMNIST\_3 & \underline{89.38} & 86.46 & 50.00 & \textbf{89.83} & 88.88 & 54.67 \\
        FashionMNIST\_9 & 94.95 & \underline{96.07} & 50.00 & \textbf{96.22} & 95.99 & 50.87 \\
        MVTec-AD\_cable & 70.56 & \underline{72.17} & 50.00 & \textbf{74.38} & 70.61 & 54.32 \\
        MVTec-AD\_leather & 99.22 & 99.39 & 50.00 & 99.38 & \underline{99.43} & \textbf{99.46} \\
        MVTec-AD\_pill & \textbf{63.21} & 61.98 & 50.00 & \underline{63.07} & 61.31 & 53.10 \\
        MNIST-C\_shear & 64.15 & 66.92 & 50.00 & \underline{67.75} & \textbf{68.73} & 52.71 \\
        CIFAR10\_7 & 63.69 & \textbf{67.06} & 58.80 & \underline{66.82} & 66.30 & 52.21 \\
        CIFAR10\_8 & 70.73 & \underline{73.13} & 50.00 & \textbf{73.64} & 73.04 & 53.42 \\
        MVTec-AD\_capsule & \textbf{66.79} & 65.01 & 50.00 & \underline{66.04} & 65.02 & 44.15 \\
        MNIST-C\_fog & 86.51 & \underline{91.09} & 50.00 & \textbf{92.60} & 90.78 & 51.15 \\
        SVHN\_2 & 61.10 & 63.68 & 50.00 & \textbf{64.20} & \underline{64.14} & 60.94 \\
        \bottomrule
    \end{tabular}
    \end{adjustbox}

    \vspace{2pt}
    \caption{Performance Comparison of Anomaly Detection Methods based on AUC-ROC (in \%) for Compact Datasets. Diffusion-based methods outperform conventional methods on all datasets. Best are in \textbf{bold} and second-best are \underline{underlined}}
    \label{table1}
\end{table}

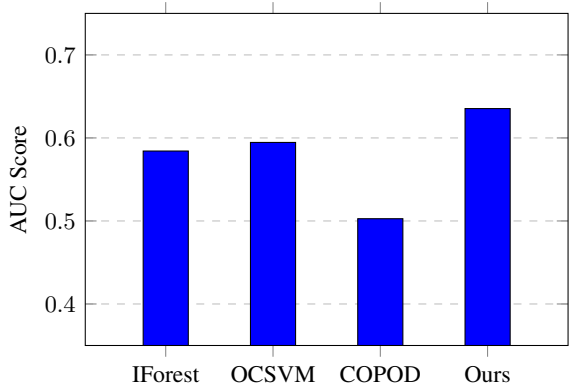
\begin{figure}[h]
    \centering
        \begin{tikzpicture}
            \begin{axis}[
                height=6cm,
                width=8cm,
                ybar=0.1pt,
                bar width=0.6cm,
                enlarge x limits=0.25,
                symbolic x coords={IForest,OCSVM, COPOD, Ours},
                xtick=data,
                ylabel={AUC Score},
                ymajorgrids=true,
                grid style=dashed,
                ymin=0.35, 
                ymax=0.75, 
                ytick={0.4, 0.5, 0.6, 0.7},
                legend image code/.code={
                \draw [#1] (0cm,0.1cm) rectangle (0.15cm,0.15cm); },
                x tick label style={rotate=0, anchor=north},
                legend style={at={(0.5,1.05)}, anchor=south, legend columns=2, nodes={scale=0.8, transform shape}},
                label style={font=\footnotesize},
                tick label style={font=\footnotesize}  
            ]
            \addplot[fill=blue] coordinates {(IForest, 0.58423398) (OCSVM, 0.594583473) (COPOD, 0.502704369) (Ours, 0.635354277)};

            \end{axis}
        \end{tikzpicture}
    \caption{AUC Scores for Anomaly Detection Methods on the High-resolution dataset (Scalability Test). The superior performance of our method shows that Diffusion Transformers are more scalable compared to conventional methods.}
    \label{miniimagenet}
\end{figure}

All experiments have been performed using AUC-ROC as the evaluation metric.
\subsubsection*{On Compact Datasets} Table \ref{table1} showcases a detailed comparison of anomaly detection methods across a variety of datasets. Overall, the Diffusion based methods consistently perform well, often surpassing or closely matching the best-performing traditional methods such as \textbf{IForest}, \textbf{OCSVM}, and \textbf{COPOD}. \textbf{DTE}, in particular, exhibits exceptional performance in certain cases, such as the MVTec-AD\_tile (92.74\%), MVTec-AD\_carpet (82.36\%), and MVTec-AD\_leather (99.46\%), demonstrating its ability to handle high-dimensional anomaly detection tasks effectively. \textbf{COPOD}, while providing a stable baseline, generally underperforms compared to other methods, with a fixed AUC score of 50.00\% on many datasets. 
\\
\subsubsection*{On High-Resolution Datasets} Plot \ref{miniimagenet} highlights the results of the scalability test on the  Mini-ImageNet dataset , a more complex and larger-scale dataset. The  AUC scores  indicate that the proposed method (Ours, AUC = 0.635) outperforms all baseline models, with  \textbf{OCSVM} (0.594)  and  \textbf{IForest} (0.584)  following behind. This observation suggests that the proposed diffusion-based method scales effectively to larger datasets while maintaining competitive performance. In contrast, traditional methods like \textbf{COPOD} demonstrate limited scalability, with significantly lower scores. 

\textbf{Robustness Across Dataset Types:} The methods' performance on  adversarial or noisy datasets (e.g., MNIST-C\_fog  and  MNIST-C\_shear ) highlights the robustness of \textbf{DDPM} and \textbf{DiT}-based methods. For instance, in the  MNIST-C\_fog  dataset, \textbf{DDPM} achieves  92.60\% , surpassing all other methods, including  OCSVM (91.09\%) .

\textbf{Scalability and Adaptability:} The scalability test using the  Mini-ImageNet dataset  provides a critical validation of the proposed method's adaptability to larger datasets with higher complexity. While traditional models (e.g., IForest and OCSVM) show moderate scalability, their AUC scores plateau, highlighting their limitations when applied to larger datasets. In contrast, the superior performance of diffusion models on this dataset indicates that advanced models like DDPM, DiT, and DTE can effectively manage increased dataset size and complexity without compromising on performance.


\section{Conclusion}
This work investigates the application of diffusion-based models for anomaly detection, demonstrating their effectiveness across a range of compact and high-resolution datasets. Diffusion-based methods, particularly DDPMs and DiTs, consistently outperform or closely match traditional methods like Isolation Forests, One-Class SVMs, and COPOD. Their ability to handle noisy and adversarial datasets further underscores their robustness. The adaptability of diffusion models to high-dimensional datasets, such as Mini-ImageNet, showcases their scalability without significant performance degradation, unlike traditional approaches. The use of reconstruction error as a metric proves critical for distinguishing anomalies, with diffusion-based architectures excelling in detecting subtle deviations.

Future research should focus on enhancing the encoder-decoder designs to better address high-dimensional challenges and exploring the integration of multi-modal datasets to expand the applicability of these methods. By establishing diffusion-based models as a strong baseline for anomaly detection, this study paves the way for advancements in adaptive and scalable detection solutions.

\bibliography{Bibliography}

\end{document}